\documentclass[manuscript]{acmart}

\renewcommand\footnotetextcopyrightpermission[1]{} 
\newcommand{\CCH}[1]{{\color{black}#1}\normalfont} 
\newcommand{\CCHR}[1]{{\color{black}#1}\normalfont}

\setcopyright{acmlicensed}
\copyrightyear{2025}
\acmYear{2025}
\acmDOI{XXXXXXX.XXXXXXX}

\acmJournal{TIST}

\usepackage{booktabs}
\usepackage{cleveref}
\usepackage{booktabs}
\usepackage{multirow}
\usepackage{graphicx} 

\usepackage{bbding}
\begin{document}

\title{ExReg: Wide-range Photo Exposure Correction via a Multi-dimensional Regressor with Attention}

\author{Huu-Phu Do}
\authornote{Equal contribution}
\affiliation{%
  \institution{National Yang Ming Chiao Tung University}
  \city{Hsinchu}
  \country{Taiwan}
}
\email{dohuuphu25.ee11@nycu.edu.tw}

\author{Hao-chien Hsueh}
\authornotemark[1]
\affiliation{%
  \institution{National Yang Ming Chiao Tung University}
  \city{Hsinchu}
  \country{Taiwan}}
\email{ck1040909.cs11@nycu.edu.tw}

\author{Tzu-hao Chiang}
\affiliation{%
 \institution{National Yang Ming Chiao Tung University}
 \city{Hsinchu}
 \country{Taiwan}}
\email{raes210109@gmail.com}

\author{Chi-han Chen}
\affiliation{%
  \institution{National Yang Ming Chiao Tung University}
  \city{Hsinchu}
  \country{Taiwan}
}
\email{ludwig1017@gmail.com}

\author{Wen-hsiao Peng}
\affiliation{%
  \institution{National Yang Ming Chiao Tung University}
  \city{Hsinchu}
  \country{Taiwan}}
\email{wpeng@cs.nctu.edu.tw}

\author{Ching-chun Huang}
\authornote{Corresponding author}
\affiliation{%
  \institution{National Yang Ming Chiao Tung University}
  \city{Hsinchu}
  \country{Taiwan}}
\email{chingchun@nycu.edu.tw}

\renewcommand{\shortauthors}{Huu-Phu Do, Hao-Chien Hsueh et al.}

\begin{abstract}
  Photo exposure correction is widely investigated, but fewer studies focus on correcting under- and over-exposed images simultaneously. Three issues remain open to handle and correct both under- and over-exposed images in a unified way. First, a locally-adaptive exposure adjustment may be more flexible instead of learning a global mapping. Second, it is an ill-posed problem to determine the suitable exposure values locally. Third, photos with the same content but different exposures may not reach consistent adjustment results. To this end, we proposed a novel exposure correction network, ExReg, to address the challenges by formulating exposure correction as a multi-dimensional regression process. Given an input image, a compact multi-exposure generation network is introduced to generate images with different exposure conditions for multi-dimensional regression and exposure correction in the next stage. An auxiliary module is designed to predict the region-wise exposure values, guiding the proposed Encoder-Decoder ANP (Attentive Neural Processes) to regress the final corrected image. The experimental results show that ExReg can generate well-exposed results and outperform the SOTA method in PSNR for extensive exposure problems. Furthermore, the processing speed, with 0.05 seconds per image on an RTX 3090, is efficient. When tested on the same image under various exposure levels, ExReg also yields results that are visually consistent and physically accurate.
\end{abstract}



\maketitle

\section{Introduction}


Photos have been popular mobile multimedia ever since the spread of portable devices and social media. However, improper exposure time settings would degrade the visual quality, lead to unpleasant luminance, and even affect the subsequent computer vision tasks such as human detection and recognition. Over the past few decades, some studies have investigated the enhancement of wrongly exposed images; the related works can be divided into two groups. The first category applies an unsupervised scheme to train the network with unpaired data and builds its architecture upon GAN to map the distribution of low-quality images to that of high-quality ones~\cite{ni2020towards},~\cite{jiang2021enlightengan}. However, these GAN-based methods suffer the issue of color deviation and tend to generate fake content. \CCHR{Recently, EnlightenGAN \cite{jiang2021enlightengan} has introduced the use of perceptual loss between the input and enhanced images to mitigate artifacts. However, this approach also imposes a constraint that limits the model's ability to effectively handle a wide range of lighting conditions.} The other category, employing CNNs and relying on paired data for supervised training, results in highly database-dependent performance. For example, the models trained on a database containing more under-exposed images~\cite{Chen2018Retinex} tend to over-enhance the images and vice versa. To extend the ability to tackle a variety of lighting conditions in a unified manner,~\cite{afifi2021learning} targets both the under-exposed and over-exposed images by generating a new dataset with abundant lighting conditions. They then proposed a coarse-to-fine architecture consisting of 4 U-Nets to correct variant exposure errors via the rich training resources. However, the abovementioned methods implicitly look for a corrected image under an ideal and global exposure. If a testing image contains several luminance conditions in different regions, the related works may fail to enhance it regionally. To this end, we proposed a novel photo exposure correction framework based on multi-dimensional regression that can deal with diverse exposure errors and perform image enhancement in a local and adaptive manner.

\begin{figure}[hb] 
\centering 
\includegraphics[width=0.45\textwidth]{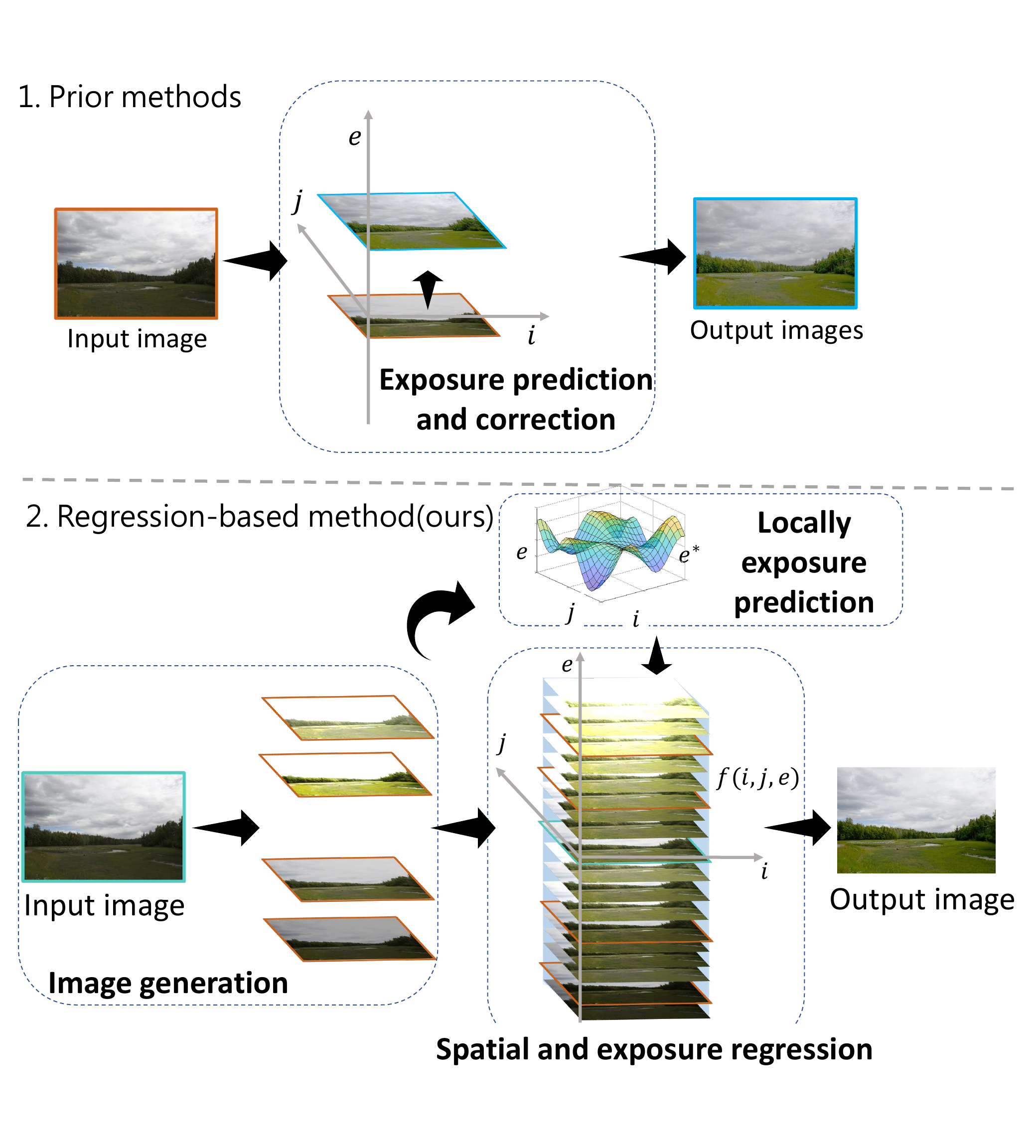} 
\caption{Concept comparison between \textbf{(1)} Conventional Methods Using Global Exposure Correction and \textbf{(2)} Our Regression-Based Method: Traditional methods typically rely on implicitly estimating the \textbf{global} exposure level and subsequently applying corrections via a learning network. In comparison, our work initially estimates the local continuous exposure values (top-right) and produces the enhanced image by referring to regressed multi-dimensional function $f(i,j,e)$ given the target location-exposure indexes $(i,j,e)$ (bottom-right).}
\label{fig:priorvsour}
\end{figure}


Lately, attentive neural processes (ANP)~\cite{kim2019attentive} proposed to regress the intensity value of a pixel as a function of its location. It applied self-attention to extract multiple representations from the input pairs (coupling coordinates and their corresponding intensity) and introduced cross-attention among the learned representations and the query followed by a multilayer perceptron (MLP) to regress the output intensity continuously. In this way, ANP utilizes self-attention to depict a family of conditional distributions for regression (behaving as Gaussian processes) and leverages the power of cross-attention to prevent prediction underfitting. Inspired by the continuous representation of ANP, we proposed a novel network, ExReg, which aims to regard exposure correction as a multi-dimensional regression problem. As shown in the bottom of Fig.~\ref{fig:priorvsour}, to provide a fine exposure correction up to a continuous level locally, our correction process dynamically modeled a regression function $f(i,j,e)$ that maps location-and-exposure coordinates to pixel-wise RGB values. Here, given an input image, the other four images under different exposure values are automatically generated. Next, we use the five images to construct $f(i,j,e)$ for the input image via the proposed Encoder-Decoder ANP. After involving an auxiliary network module to estimate the correct exposure value $e_{i,j}^\ast$ for each image location $(i,j)$, we finally obtain a well-exposed image by searching the manifold surface cutting from the regression space according to $(i,j,e_{i,j}^\ast)$. Compared with the conventional works that correct the image exposure globally, our ExReg enables the exposure correction locally and continuously.

\begin{figure*}[ht] 
\centering 
\includegraphics[width=0.95\textwidth]{figures/intro_2.pdf} 
\caption{Overview of the ExReg Framework. This diagram illustrates the ExReg framework, where MEGNet produces four distinct images under varying exposure conditions for function regression in the next stage. Subsequently, given the input image and the four generated images, RegNet is proposed to predict region-wise exposures $E^{\ast}$ and build the implicit RGB regression function over the spatial-exposure (multi-dimensional) domain via a unique encoder-ANP-decoder framework with cross attention. Here, ANP means Attentive Neural Processes.}
\label{fig:archall}
\end{figure*}

As shown in Fig. \ref{fig:archall} specifically, our ExReg decomposes the exposure correction process into two stages. In the first stage, we introduce a multi-exposure generation network (MEGNet) which consists of a base generation network and a conditional branch to produce four images with different exposure conditions from the input image ($I_{e_0}$) for regression in the next stage. In the second stage, the exposure regression network, RegNet, is proposed for region-wise exposure prediction and RGB regression over the spatial-exposure (multi-dimensional) domain using the input image and four generated images. The regression part, realized by cross-attention modules, has a unique encoder-ANP-decoder framework. Note that the original ANP utilizes the inherent pixel tokens and self-attention to achieve feature interaction; however, the high computational cost and unaffordable memory usage make ANP hard to handle large images (i.e., the number of pixel tokens is the critical issue). To address the problem, our encoder-ANP-decoder modifies the self-attention step with a convolution-based encoder to extract semantic features as regional tokens. Then, cross-attention and neural processes are designed to model the multi-dimensional regression function in the feature domain. Later, given the estimated regional exposure values, the corrected regional features are retrieved from the regression function. Finally, the corrected output image is acquired through a deconvolution-based decoder. 

Though our encoder-ANP-decoder provides a computation-efficient solution for exposure correction, the region-based processing may make the exposure-corrected image blurry and blocky. To this end, we further incorporate a lightweight Feature Adjustment Module (FAM) to preserve the original image details by introducing a skip connection path between the encoder and the decoder. On the other hand, to realize the idea of using implicit neural representation to model the multi-dimensional regression in RegNet, we design an auxiliary exposure predictor to estimate the spatially-variant exposure values $e^{\ast}$. Here, we predict $e^{\ast}$ by referring to the input image and its four MEGNet-generated images coupled with their corresponding exposure settings. In brief, our contributions can be summarized as follows:
\begin{itemize}
\item We introduce a novel photo exposure correction framework (ExReg) that incorporates image generation and multi-dimensional regression to adjust image exposure over regions.
\item 
We propose an image generation network, MEGNet, which comprises a straightforward yet effective base network coupled with a conditional branch. This design enables MEGNet to generate images with arbitrary exposure settings, thereby facilitating a more versatile approach for regression-based image enhancement.  
\item We introduce RegNet, an ANP-based exposure correction network. RegNet is designed to model regression over the spatial-exposure domain, given a limited set of images captured under various exposure settings. It functions by first estimating exposure values for the input images and then retrieves the regression model to output well-exposed images. This approach allows RegNet to effectively adjust the exposure of images, aligning them with desired exposure levels.
\end{itemize}

\section{Related Works}

\CCH{
\subsection{Exposure Correction}


There are many studies focus on correcting the exposure of sRGB images. These papers can be divided into two categories: traditional methods~\cite{he, lime, he1, he2}, and learning-based methods~\cite{gharbi2017deep, Chen2018Retinex, Wang_2019_CVPR, zhang2019kindling, denoise, jiang2021enlightengan, ni2020towards, afifi2021learning, TCSVT2, TCSVT3, TCSVT4, Cui_2022_BMVC, wang2022lcdp, hue2023psenet, ma2022toward}. In this section, we will begin by discussing traditional methods and then proceed to explore learning-based methods.

Conventionally, there are two types of traditional methods for exposure correction. One is based on histogram equalization~\cite{he}, and the other~\cite{lime} is built upon the Retinex theory (i.e., image decomposition and then enhancement). Histogram equalization performs enhancement globally by redistributing the gray levels of the whole image to achieve a more uniform distribution. Its succeeding works~\cite{he1}\cite{he2} further improve the performance by allowing local and adaptive adjustment. Retinex-based methods assume that color images can be decomposed into reflectance and illumination components. Next, brightness and exposure correction is performed on the illumination maps. 

With the rapid development of deep learning, deep Retinex-based methods have also been proposed to reach more robust decomposition and involve less handcrafted tunings~\cite{Chen2018Retinex}\cite{ Wang_2019_CVPR}. Moreover, the works~\cite{Chen2018Retinex}\cite{ zhang2019kindling} propose to enhance both the illuminance and reflectance components, which corrects the exposure and alleviates the image degradation at the same time. Later, \CCHR{Zhao et al.~\cite{TCSVT4} propose a zero-shot method using a generative strategy to weaken the coupling relationship between the two components while performing Retinex decomposition.} 

Instead of Retinex decomposition, Ke Xu et al.~\cite{denoise} decompose the input image into low-frequency and high-frequency parts. In this way, they can perform image denoising and low-light enhancement by recovering the image content, suppressing the noise in the low-frequency component, and restoring the high-frequency details. \CCHR{Distinct from approaches based on Retinex theory, Li et al.\cite{TCSVT2} propose an end-to-end low-light enhancement network, ingeniously integrating attention layers, convolutional layers, and circulation layers to reduce the overall number of network parameters. Meanwhile, Fan et al.\cite{TCSVT3} introduce a multiscale enhancement network that applies illumination constraints, effectively addressing issues of uneven exposure and color distortion.}

Besides the aforementioned supervised methods, unsupervised learning-based methods are also famous for image enhancement. The conventional approach is to apply a generative adversarial network (GAN) to learn the distribution mapping from low-quality images to high-quality ones. 
These GAN-based methods can generate normal light images with higher perceptual quality. However, GAN-based methods suffer from the issue of color deviation and tend to generate fake content. \CCHR{While EnlightenGAN~\cite{jiang2021enlightengan} introduces a perceptual loss in high-level features between the input and enhanced images to improve the fidelity of the generated images, it continues to face challenges in addressing color deviation and accommodating a broad spectrum of lighting conditions.}

It is worth noting that the above methods, mainly focusing on enhancing low-light images, make them unsuitable for correcting overexposed errors. 
\CCHR{To address the issue, Afifi et al.~\cite{afifi2021learning} proposed a unified framework to enhance both over- and under-exposed images.} Inspired by Laplacian pyramid reconstruction, they also proposed a coarse-to-fine network consisting of 4 U-Nets that can enhance the color and details information sequentially and deal with variant exposure errors. Furthermore, Wang et al.~\cite{wang2022lcdp} propose a local color distribution embedded module to jointly address both over- and under-exposure problems in a single image.} However, these methods must rely on a synthesized dataset with abundant lighting conditions.

\subsection{Neural Function Representation and Regression}
Regression has been a classic problem; considerable research studies have appeared, such as Gaussian processes, CNN-based methods, function representation learning, etc. Despite the remarkable success,  conventional learning-based methods~\cite{lake2017},~\cite{garnelo2016deep} need to train a new model for a new function and require a large dataset. In contrast, Gaussian processes (GPs), which specify a distribution over functions without a costly training phase, can quickly infer a new function during testing by exploiting the prior knowledge from observations. However, it becomes computationally intractable as the number of observations grows. Recently, some works have tried to combine the idea of the Gaussian process with a neural network training regime to benefit from both sides. To name some, given a set of observations, Conditional Neural Processes (CNPs)~\cite{cnp} define conditional distributions over functions that are parameterized by a neural network. Unlike conventional supervised deep learning methods for function modeling, CNPs do not need to be re-trained for a new function. However, CNPs are proposed to capture the uncertainty at a local output level, making it challenging to model the diverse realizations of the data-generating process.

To address this issue and allow a more flexible model, Garnelo et al.~\cite{np} propose Neural Processes (NPs) and introduce a global latent variable that captures the global uncertainty. Nevertheless, NPs tend to underfit the observations due to the mean-aggregation step after the encoder. Taking an average across the representations of context observations gives the same weight to each context point. However,
Different context points should contribute differently to the predicted target. In GPs, the property is realized through the concept of kernels, while in neural networks, the differentiable cross-attention mechanism can perform a similar role. Based on this idea, Kim et al. ~\cite{kim2019attentive} propose Attentional Neural Processes (ANPs) that can address the underfitting issue by incorporating cross-attention and self-attention with NPs. They show that ANPs can give a more accurate predictive mean and a smaller variance than NPs in the 1D function regression problem. Regarding 2D function regression like image data, NPs still suffer a large variance and fail to give accurate inference results; in contrast, ANPs can generate images much closer to the ground truth. Consequently, ANPs are more suitable to be adopted for image processing tasks.

\section{Proposed Method}\label{sec:method}
\CCH{
As shown in \Cref{fig:archall}, our ExReg consists of two components: multi-exposure generation network (MEGNet) and exposure regression network (RegNet). First, given an RGB input image $I$ with an arbitrary exposure setting, MEGNet generates four images; two of them are less exposed while the others are more exposed than the input image. With the four derived images and the input one, RegNet predicts the correct exposure values over the spatial domain and produces the output image $Y$ by attention-based regression. The details of each component are elaborated on in the following subsections. \CCHR{For better readability, the definition of the notations used in the proposed method are summarized in \Cref{tab:notation}.} 
}

\begin{table}[t!]
\centering
\caption{Definition of the notations.}
\begin{tabular}{c|c}
\toprule
Notation & Meaning \\
\midrule
$I_{e_0}$ & Input image \\
$I_{e_0^{--}}$, $I_{e_0^{-}}$ & Less exposed image generated by MEGNet \\
$I_{e_0^{++}}$, $I_{e_0^{+}}$ & More exposed image generated by MEGNet \\
$F_{e}$ & Extracted features from $I_{e}$ \\
$F^\ast_{e^\ast}$ & Enhanced corrected features \\
$f_{ije}$ &  Features of  exposure value ($e$) in position $(i,j)$ \\
$f'_{ije_{ij}^\ast}$ & First step corrected features in position $(i,j)$  \\
$f_{ije_{ij}^\ast}^\ast$ & Second step enhanced corrected features  in position $(i,j)$ \\ 
$f_{n}$ & Intermediate features at the $n^{th}$ layer \\
$E^\ast$ & Pixel-wise optimized exposure map \\
$e_{ij}^\ast$ &  Optimized exposure value in position $(i,j)$\\

\bottomrule

\end{tabular}
\label{tab:notation}
\end{table}

\subsection{MEGNet}

CSRNet~\cite{he2020conditional} shows that $1\times1$ convolutional operations can realize global brightness changes and contrast adjustment. Due to its simplicity and effectiveness, our MEGNet also uses $1\times1$ convolution to constitute our generator and produce images under different exposure values. Furthermore, to generate images according to the given exposure, we design a conditional branch inspired by TMNet~\cite{xu2021temporal}. The exposure difference$\Delta EV$ between the output and the input images is transformed to the exposure condition by fully connected layers to denote the lighting difference between the two images. Moreover, since the generation is manipulated pixel by pixel independently through $1\times1$ convolution, a global image prior is also incorporated in the conditional branch to guide the process and ensure global image consistency. 

Based on MEGNet, we generate four images, denoted as $I_{e_0^{--}}, I_{e_0^-},  I_{e_0^+}, I_{e_0^{++}}$, by referring the input image $I_{e_0}$ and four corresponding $\Delta EV=-1.5,-1.0,1.0,1.5$. The base generation network involves three $1\times1$ convolutional layers with the activation function ReLU. The global brightness information is extracted by an encoder and concatenated with the exposure-related condition to generate the feature modulation parameters by two fully connected layers. Given the parameters, the conditional branch dynamic modulates the intermediate features within the base network by performing the following affine transformation:
\begin{equation}
    MEGf\:'_{n} = \alpha_{n} \cdot MEGf_{n} + \beta_{n},
\end{equation}
where $MEGf_{n}$ is the intermediate feature at the $n^{th}$ layers ($n=1\sim3$); $\alpha_{n}$ and $\beta_{n}$ are the modulation parameters.
\begin{figure*}[t] 
\centering
\includegraphics[width=0.75\textwidth]{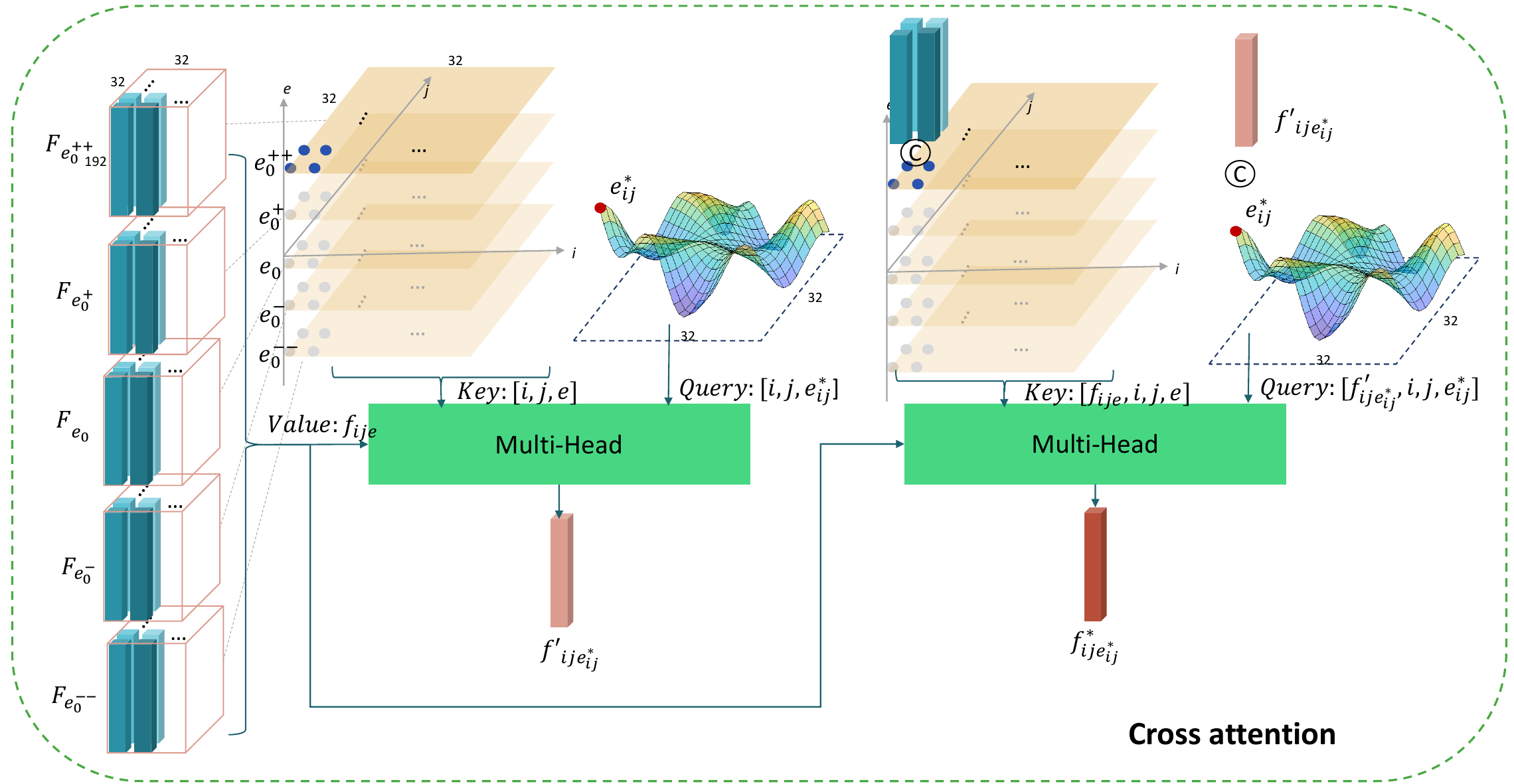} 
\caption{Cross-Attention Module Functionality - This figure illustrates the two-stage operation of the cross-attention module for feature regression and enhancement. In the first stage, the module computes the preliminary corrected features, $f'_{ije_{ij}^\ast}$, corresponding to the estimated optimal exposure values $e_{ij}^\ast$ at their respective locations $(i,j)$. The second stage employs feature attention, utilizing the feature vectors as keys and queries, to further refine and produce the enhanced features, $f_{ije_{ij}^\ast}^\ast$, for each unique location $(i,j)$.}
\label{fig:cross}
\end{figure*}


\subsection{RegNet}
\label{sec:method_RegNet}
In our RegNet, as depicted in \Cref{fig:archall}, an optimal exposure map is first estimated; then, an illumination-corrected image is generated through the proposed multi-dimensional regression. Inspired by ANP~\cite{kim2019attentive}, the regression part is composed of an encoder, a cross-attention-based regressor, and a decoder part. As mentioned earlier, to achieve the locally adaptive correction and avoid heavy computational load, we replace the self-attention-based encoder in ANP with a convolutional one. Given the input image with its size $H$ (i.e., 512) and $W$ (i.e., 512), the images, $I_{e_0^{--}}, I_{e_0^-}, I_{e_0},  I_{e_0^+}, I_{e_0^{++}} \CCHR{(\in\mathcal{R}^{H\times W\times3})}$, will be fed into the encoder to obtain five feature maps $F_{e_0^{--}},F_{e_0^-},F_{e_0},F_{e_0^+},F_{e_0^{++}}\CCHR{(\in\mathcal{R}^{\frac{H}{16}\times\frac{W}{16}\times192})}$. Let $f_{ije}$ denote $F_e$'s $ij$-th element, where $(i,j)$ spans the spatial dimensions, and $e \in \{ e_0^{--},e_0^-,e_0,e_0^+,e_0^{++} \}$. In our cross-attention module, $f_{ije}$ is treated as the value, and the corresponding key is the concatenation of $i,j,$ and $e$ (i.e., that is $[i,j,e]$). 


Moreover, to predict a suitable pixel-wise exposure map $E^\ast$, $I_{e_0^{--}}, I_{e_0^-}, I_{e_0},  I_{e_0^+}, I_{e_0^{++}}$ is further concatenated with their exposure value $e_0^{--}, e_0^-, e_0,  e_0^+, e_0^{++}$ as input of our single-scale exposure prediction module, as shown in \Cref{fig:archall}. To approximate the correct exposure value for each feature token (i.e., its receptive field covers a local region), we use an averaging pooling to downsample $E^\ast$ to \CCHR{$\frac{H}{16}\times\frac{W}{16}$} so that its $(i,j)^{th}$ element is the corresponding exposure value $e_{ij}^\ast$ for a feature token $f_{ije}$. Compared with encoder-decoder architecture, the proposed single-scale pipeline can preserve more exposure details because it does not contain a downsampling operation. The spatial size of $E^\ast$ is the same as the input image so that we can reuse $E^\ast$ for our feature adjustment module.

After acquiring the optimized/estimated exposure values $e_{ij}^\ast$ for all $(i,j)$, we treat the concatenation $[i,j,e_{ij}^\ast]$ as the query information used in the cross-attention step. The two-stage corrected feature $f_{ije_{ij}^\ast}^\ast$ is then predicted by the cross-attention module that contains two 8-head attention blocks. Specifically, our idea is to split the regression into two steps (corresponding to the two attention blocks). The first block/step is to interpolate the preliminary corrected features $f'_{ije_{ij}^\ast}$ corresponding to the estimated exposure value and its locations, while the second block/step is to focus on obtaining visually pleasing tones by involving the feature vectors in the keys and queries to obtain feature-attentive results. As shown in \Cref{fig:cross}, the two-stage corrected feature $f_{ije_{ij}^\ast}^\ast$ given its exposure value $e_{ij}^\ast$ can be obtained by our cross-attention based multi-dimensional regressor. Following up, we design a decoder to map the corrected features $f_{ije_{ij}^\ast}^\ast$, where $i,j\in\{ 1,2,...,32\}$, back to the output RGB image, $Y$.

However, the reconstructed image tends to be blurry due to the information loss caused by the convolutional encoder and decoder. To deal with this problem, a skip connection path is constructed between the encoder and decoder to propagate the detailed information. A feature adjustment module (FAM) is introduced in the path to adjust the encoded features from the input image $I_{e_0}$ before merging them to the decoder. Let $ENf_n(\in\mathcal{R}^{H\times W \times C})$ be the encoded features of $I_{e_0}$ in the $n^{th}$ layer and $E^\ast\CCHR{(\in\mathcal{R}^{H\times W})}$ be the estimated exposure map of $I_{e_0}$ used for exposure correction. The FAM takes $ENf_n$ and $E^\ast$ as inputs, where $E^\ast$ is resized to $H \times W$ to match the size of $ENf_n$. The resized $E^\ast$ are used to predict a scaling map $\mathcal{S}_n(\in \mathcal{R}^{H\times W})$ and a bias map $\mathcal{B}_n(\in \mathcal{R}^{H\times W})$ separately via two different $1\times1$ convolutional networks. In this way, the exposure-modulated feature of the $n^{th}$ layer, denoted as $FAMf_n = {S}_n \cdot ENf_n + {B}_n$, can preserve more exposure-corrected details for image reconstruction using the decoder.

\subsection{Losses and Training Details}
\label{sec:LossesTraining}

The MEGNet and RegNet are trained separately at the beginning using the dataset proposed by ~\cite{afifi2021learning}, referred to as the MSEC dataset. The MSEC dataset is rendered from the MIT-Adobe FiveK dataset~\cite{mit5k} using Adobe Camera Raw SDK. Regarding the exposure value (EV=$e$) of the original input images ($I_e$) as the reference, M. Afifi et al.~\cite{afifi2021learning} use the relative EVs (i.e., $\Delta EV \in ({-1.5, -1, +1, +1.5)}$ to render images (i.e., $I_{e+\Delta EV}$) with underexposure settings (-1.5 and -1) and overexposure ones (+1 and +1.5). Also, they adopt the images manually retouched by Expert C in MIT-Adobe FiveK dataset as the correctly exposed images (i.e., the ground truth $Y_{gt}$).

To train MEGNet for image generation, we collect training pairs from the dataset. In detail, for an input image $I_e$ whose EV is $e$, we randomly draw the  $\Delta EV$ from ${-1.5, -1, +0, +1, +1.5}$ and select its corresponding image $I_{e_{gt}}$ as the output, where $e_{gt}=e+\Delta EV$. Then, given the input image $I_e$ and $\Delta EV$, we train MEGNet to generate the image $I_{out}$ that is constrained to be close to $I_{e_{gt}}$ by L1 reconstruction loss :
\begin{equation}
    \mathcal{L}_{MEG} = \sum_{\ training\ set} ||I_{out}-I_{e_{gt}}||_1.
\end{equation}

After training MEGNet, we fixed its parameters and trained the RegNet. Given each (under or over-exposed) image $I$ in the training dataset, RegNet is trained to generate the enhanced image $Y$ under the following Charbonnier loss:

\begin{equation}
    \mathcal{L}_{Reg} = \sum_{training\ set} \sqrt{(Y-Y_{gt})^{2}+\epsilon^2},
\end{equation}
where $Y_{gt}$ is the retouched ground-truth image and $\epsilon=10^{-3}$. Later, we co-trained MEGNet and RegNet in an end-to-end manner \CCHR{using the Charbonnier loss mentioned above} to achieve the final exposure correction model, ExReg. 

\section{Experiments}
\CCH{
In this section, we first validate the effectiveness of our proposed ExReg framework by comparing it against a broad set of existing methods for exposure correction and image enhancement in Sec.~\ref{sec:ExReg}. We then conduct detailed ablation studies to assess the contribution of each component in our model, as presented in Sec.~\ref{sec:AblationRegNet} and Sec.~\ref{sec:ExpRegNet}. For clarity, we refer to the architecture proposed by M. Afifi et al.~\cite{afifi2021learning} as MSEC throughout this paper.
}

\subsection{ExReg}
\label{sec:ExReg}
\CCH{
\paragraph{Datasets} We train our ExReg model using the training set described in Sec.~\ref{sec:LossesTraining}, which consists of 17,675 training images and 5,905 testing images. For evaluation, we use the testing set comprising 3,543 well-exposed and over-exposed images (EVs = +0, +1.0, +1.5), and 2,362 under-exposed images (EVs = -1, -1.5). To further assess the generalization capability of our model, we conduct evaluations on several external datasets that span a wide range of real-world lighting conditions---LIME~\cite{lime}, NPE~\cite{npe}, VV~\cite{vv}, and DICM~\cite{dicm}---as well as on challenging nighttime and low-light benchmarks including DarkFace~\cite{darkface}, MEF~\cite{mef}, and LOL~\cite{lol}. These datasets are used exclusively for testing and are not involved in the training process.

}
\paragraph{Implementation detail} The base network of MEGNet contains 3 convolutional layers with channel size 64 and kernel size $1\times1$ while its conditional branch contains a global information encoder with three convolutional layers whose channel size is 32 and the other exposure condition encoder with 3 MLPs layers. On the other hand, the architecture of RegNet is composed of an Encoder-Decoder structure with a 4-layer  $3 \times 3$  convolutional network, a cross-attention module with two 8-head multi-head attention blocks, and an exposure predictor realized by a 7-layer single-scale convolutional network. In total, our ExReg, including both MEGNet and RegNet, has 2.68M parameters. The training process of ExReg is split into two stages. First, we pre-train the MEGNet with a batch size of 100 and a patch size of $256\times256$. Then, the RegNet is trained with a batch size of 12 and a patch size of $512\times512$. We train the MEGNet for 100 epochs and train the RegNet for 200 epochs with the learning rate initialized as $1\times10^{-4}$. Later, we co-train MEGNet and RegNet \CCHR{using a batch size of 6 and a patch size of $512\times512$ for 10 epochs.} Adam optimizer is used to minimize the loss function. The computational resources required for training and testing our proposed ExReg model are detailed in \Cref{tab:m_complexity}. All evaluations were performed using Nvidia RTX 3090 GPUs. \CCHR{During the inference, we take the full size of the image as input, then compare the output image with the ground truth for PSNR and SSIM evaluation.}

\begin{table}[tp]
\centering
\caption{{Model complexity of MEGNet and RegNet}}
\scalebox{1}{
\begin{tabular}{crrrrrr}
\toprule
   &  Trainable & Training  & \multicolumn{2}{r}{Memory footprint} & Inference  & FLOPs \\
      &  parameters &  time & Training & Testing &   time&   \\
\midrule
MEGNet             & 47.05 k & 1.6 hrs & 13 Gib & 2.2 Gib & 0.002 s & 1.73 GMac\\
RegNet               & 2.64 M & 60 hrs & $2\times22$ Gib & 2.5 Gib & 0.0048 s & 69.69 GMac \\

\bottomrule
\end{tabular}
}
\label{tab:m_complexity}
\end{table}

\begin{table}[!t]
\centering
\caption{Quantitative comparison. The best results are indicated in boldface type, and the second-best ones are marked by underlines. Note that \CCHR{$^\ast$ indicates the model is trained on MSEC~\cite{afifi2021learning} training set.}}
\scalebox{0.76}{
{
\begin{tabular}{l|cc|cc|cc|cc|cc|cc}
\toprule
\multicolumn{1}{l|}{Method}  &  \multicolumn{2}{c|}{Expert A} & \multicolumn{2}{c|}{Expert B} & \multicolumn{2}{c|}{Expert C} &  \multicolumn{2}{c|}{Expert D} & \multicolumn{2}{c|}{Expert E} & \multicolumn{2}{c}{Avg.} \\
\midrule
\multicolumn{1}{l|}{} & PSNR & SSIM & PSNR & SSIM & PSNR & SSIM & PSNR & SSIM & PSNR & SSIM & PSNR & SSIM \\
\multicolumn{13}{c}{+0, +1, and +1.5 relative EVs (3543 properly exposed and over-exposed images)} \\
\midrule
HE~\cite{he} (2001) & 16.14 & 0.686 & 16.28 & 0.672 & 16.53 & 0.699 & 16.64 & 0.699 & 17.32 & 0.691 & 16.58 & 0.683 \\
HDR CNN~\cite{hdrcnn} (2017) w/PS & 14.80 & 0.651 & 15.62 & 0.689 & 15.35 & 0.670 & 16.58 & 0.685 & 18.02 & 0.703 & 16.08 & 0.680 \\
DPED (Sony)~\cite{ignatov2017dslr} (2017) & 16.40 & 0.672 & 17.68 & 0.707 & 17.38 & 0.697 & 18.00 & 0.685 & 18.69 & 0.700 & 17.63 & 0.692 \\
DPE (S-FiveK)~\cite{Chen:2018:DPE} (2018) & 14.79 & 0.638 & 15.52 & 0.649 & 15.63 & 0.668 & 16.59 & 0.664 & 17.66 & 0.684 & 16.04 & 0.661 \\
HQEC~\cite{HQEC} (2018) & 11.78 & 0.607 & 12.54 & 0.631 & 12.13 & 0.627 & 13.42 & 0.652 & 14.51 & 0.675 & 12.88 & 0.638 \\
Deep UPE~\cite{Wang_2019_CVPR} (2019) & 10.05 & 0.532 & 10.46 & 0.568 & 10.31 & 0.557 & 11.58 & 0.591 & 12.64 & 0.619 & 11.01 & 0.573 \\
Zero-DCE~\cite{Zero-DCE} (2020) & 10.12 & 0.503 & 10.77 & 0.502 & 10.40 & 0.514 & 11.47 & 0.522 & 12.35 & 0.557 & 11.02 & 0.520 \\
UEGAN~\cite{ni2020towards} (2020) & 15.02 & 0.716 & 15.98 & 0.779 & 15.74 & 0.758 & 16.64 & 0.773 & 17.35 & 0.785 & 16.15 & 0.732 \\
MSEC$^\ast$~\cite{afifi2021learning} (2021) & 18.98 & 0.743 & 19.77 & 0.731 & 19.98 & 0.768 & 18.97 & 0.716 & 19.06 & 0.727 & 19.35 & 0.737 \\
SCI~\cite{ma2022toward} (2022) & 09.41 & 0.580 & 09.87 & 0.642 & 09.66 & 0.618 & 10.79 & 0.649 & 11.74 & 0.691 & 10.29 & 0.636 \\
IAT$^\ast$~\cite{Cui_2022_BMVC} (2022) & 19.50 & {0.796} & 20.90 & 0.834 & 21.20 & 0.842 & 19.70 & 0.822 & 19.60 & 0.820 & 20.18 & 0.823 \\ 
LCDPNet$^\ast$~\cite{wang2022lcdp} (2022) & \underline{20.54} & \textbf{0.816} & \underline{21.97} & \textbf{0.849} & \underline{22.34} & \underline{0.861} & \underline{20.35} & \underline{0.830} & \underline{20.36} & \underline{0.836} & \underline{21.11} & \underline{0.838} \\
PSENet$^\ast$~\cite{hue2023psenet} (2023) & 17.89 & 0.767 & 19.79 & \underline{0.846} & 18.92 & 0.802 & 19.43 & 0.815 & 19.77 & \underline{0.836} & 19.16 & 0.813 \\
\midrule
Ours$^\ast$ & \textbf{20.68} & \underline{0.815} & \textbf{22.33} & \textbf{0.849} & \textbf{22.89} & \textbf{0.865} & \textbf{20.49} & \textbf{0.832} & \textbf{20.56} & \textbf{0.844} & \textbf{21.39} & \textbf{0.841} \\
\midrule
\multicolumn{13}{c}{-1 and -1.5 relative EVs (2362 under-exposed images)} \\
\midrule
HE~\cite{he} (2001) & 16.16 & 0.683 & 16.29 & 0.669 & 16.52 & 0.692 & 16.63 & 0.665 & 17.28 & 0.684 & 16.58 & 0.679 \\
HDR CNN~\cite{hdrcnn} (2017) w/PS & 17.32 & 0.692 & 18.99 & 0.714 & 18.05 & 0.696 & 18.38 & 0.689 & 19.59 & 0.701 & 18.47 & 0.698 \\
DPED (Sony)~\cite{ignatov2017dslr} (2017) & 18.95 & 0.679 & 20.07 & 0.691 & 18.98 & 0.662 & 17.45 & 0.629 & 15.86 & 0.601 & 18.26 & 0.652 \\
DPE (S-FiveK)~\cite{Chen:2018:DPE} (2018) & {20.15} & 0.738 & 20.97 & 0.697 & {20.92} & 0.738 & 19.05 & 0.688 & 17.51 & 0.648 & 19.72 & 0.702 \\
HQEC~\cite{HQEC} (2018) & 15.80 & 0.692 & 17.37 & 0.718 & 16.59 & 0.700 & 17.09 & 0.705 & 17.68 & 0.716 & 16.91 & 0.706 \\
Deep UPE~\cite{Wang_2019_CVPR} (2019) & 17.83 & 0.728 & 19.06 & 0.754 & 18.76 & 0.745 & 19.64 & {0.737} & \underline{20.24} & {0.740} & 19.11 & 0.741 \\
Zero-DCE~\cite{Zero-DCE} (2020) & 13.94 & 0.585 & 15.24 & 0.593 & 14.55 & 0.589 & 15.20 & 0.587 & 15.89 & 0.614 & 14.96 & 0.594 \\
UEGAN~\cite{ni2020towards} (2020) & 18.38 & 0.734 & 18.87 & 0.738 & 18.61 & 0.748 & 17.17 & 0.722 & 15.88 & 0.682 & 17.78 & 0.725\\
MSEC$^\ast$~\cite{afifi2021learning} (2021) & 19.43 & 0.750 & 20.59 & 0.739 & 20.54 & {0.770} & 18.99 & 0.723 & 18.87 & 0.727 & 19.69 & {0.742} \\
SCI~\cite{ma2022toward} (2022) & 17.90 & 0.766 & 19.52 & {0.846} & 18.80 & 0.799 & 19.30 & 0.813 & 19.47 & \underline{0.820} & 19.00 & 0.809 \\
IAT$^\ast$~\cite{Cui_2022_BMVC} (2022) & 19.82 & \underline{0.778} & 21.59 & 0.812 & 21.34 & 0.814 & 19.55 & 0.801 & 18.93 & 0.787 & 20.25 & 0.798 \\ 
LCDPNet$^\ast$~\cite{wang2022lcdp} (2022) & \underline{20.64} & \textbf{0.810} & \underline{21.95} & 0.838 & \underline{22.24} & \underline{0.846} & \underline{20.04} & \underline{0.817} & 19.64 & 0.813 & \underline{20.90} & \underline{0.825} \\
PSENet$^\ast$~\cite{hue2023psenet} (2023) & 18.62 & 0.774 & 20.80 & \textbf{0.856} & 19.36 & 0.796 & 18.96 & 0.804 & 18.71 & {0.814} & 19.29 & 0.809 \\

\midrule
Ours$^\ast$ & \textbf{20.68} & \textbf{0.810} & \textbf{22.33} & \underline{0.847} & \textbf{22.70} & \textbf{0.852} & \textbf{20.36} & \textbf{0.825} & \textbf{20.43} & \textbf{0.830} & \textbf{21.30} & \textbf{0.833} \\
\midrule
\multicolumn{13}{c}{The average performance of over- and under-exposed images (5,905 images)} \\
\midrule
HE~\cite{he} (2001) & 16.15 & 0.685 & 16.28 & 0.671 & 16.53 & 0.696 & 16.64 & 0.668 & 17.31 & 0.688 & 16.58 & 0.682 \\
HDR CNN~\cite{hdrcnn} (2017) w/PS & 15.81 & 0.667 & 16.97 & 0.699 & 16.43 & 0.681 & 17.30 & 0.687 & 18.65 & 0.702 & 17.03 & 0.687 \\
DPED (Sony)~\cite{ignatov2017dslr} (2017) & 17.42 & 0.675 & 18.64 & 0.701 & 18.02 & 0.683 & 17.55 & 0.660 & 17.78 & 0.663 & 17.88 & 0.676 \\
DPE (S-FiveK)~\cite{Chen:2018:DPE} (2018) & 16.93 & 0.678 & 17.70 & 0.668 & 17.74 & 0.696 & 17.57 & 0.674 & 17.60 & 0.670 & 17.51 & 0.677 \\
HQEC~\cite{HQEC} (2018) & 13.39 & 0.641 & 14.47 & 0.666 & 13.91 & 0.656 & 14.89 & 0.674 & 15.78 & 0.692 & 14.49 & 0.666 \\
Deep UPE~\cite{Wang_2019_CVPR} (2019) & 13.16 & 0.610 & 13.90 & 0.642 & 13.69 & 0.632 & 14.81 & 0.649 & 15.68 & 0.667 & 14.25 & 0.640 \\
Zero-DCE~\cite{Zero-DCE} (2020) & 11.64 & 0.536 & 12.56 & 0.539 & 12.06 & 0.544 & 12.96 & 0.548 & 13.77 & 0.580 & 12.60 & 0.549 \\
UEGAN~\cite{ni2020towards} (2020) & 16.37 & 0.723 & 17.14 & 0.763 & 16.89 & 0.754 & 16.86 & 0.753 & 16.77 & 0.744 & 16.80 & 0.747 \\
MSEC$^\ast$~\cite{afifi2021learning} (2021) & {19.16} & 0.746 & 20.10 & 0.734 & 20.21 & 0.769 & 18.98 & 0.719 & 18.98 & 0.727 & 19.48 & 0.739 \\
SCI~\cite{ma2022toward} (2022) & 12.80 & 0.654 & 13.73 & 0.723 & 13.31 & 0.690 & 14.20 & 0.715 & 14.83 & 0.743 & 13.78 & 0.705 \\

IAT$^\ast$~\cite{Cui_2022_BMVC} (2022) & 19.63 & \underline{0.789} & 21.18 & 0.825 & 21.26 & 0.831 & 19.64 & 0.813 & 19.33 & 0.807 & 20.21 & 0.813 \\ 
LCDPNet$^\ast$~\cite{wang2022lcdp} (2022) & \underline{20.58} & \textbf{0.813} & \underline{21.96} & {0.845} & \underline{22.30} & \underline{0.855} & \underline{20.23} & \underline{0.825} & \underline{20.07} & \underline{0.827} & \underline{21.03} & \underline{0.833} \\ 
PSENet$^\ast$~\cite{hue2023psenet} (2023) & 18.18 & 0.770 & 20.19 & \textbf{0.850} & 19.09 & 0.800 & 19.24 & 0.811 & 19.35 & \underline{0.827} & 19.21 & 0.812 \\ 

\midrule
Ours$^\ast$ & \textbf{20.68} & \textbf{0.813} & \textbf{22.33} & \underline{0.848} & \textbf{22.81} & \textbf{0.860} & \textbf{20.44} & \textbf{0.829} & \textbf{20.51} & \textbf{0.838} & \textbf{21.35} & \textbf{0.838} \\ 
\bottomrule
\end{tabular}
}
}

\label{tab:quantresult}
\end{table}

\begin{figure*}[] 
\centering
\scalebox{0.93}{
\includegraphics[width=0.95\textwidth]{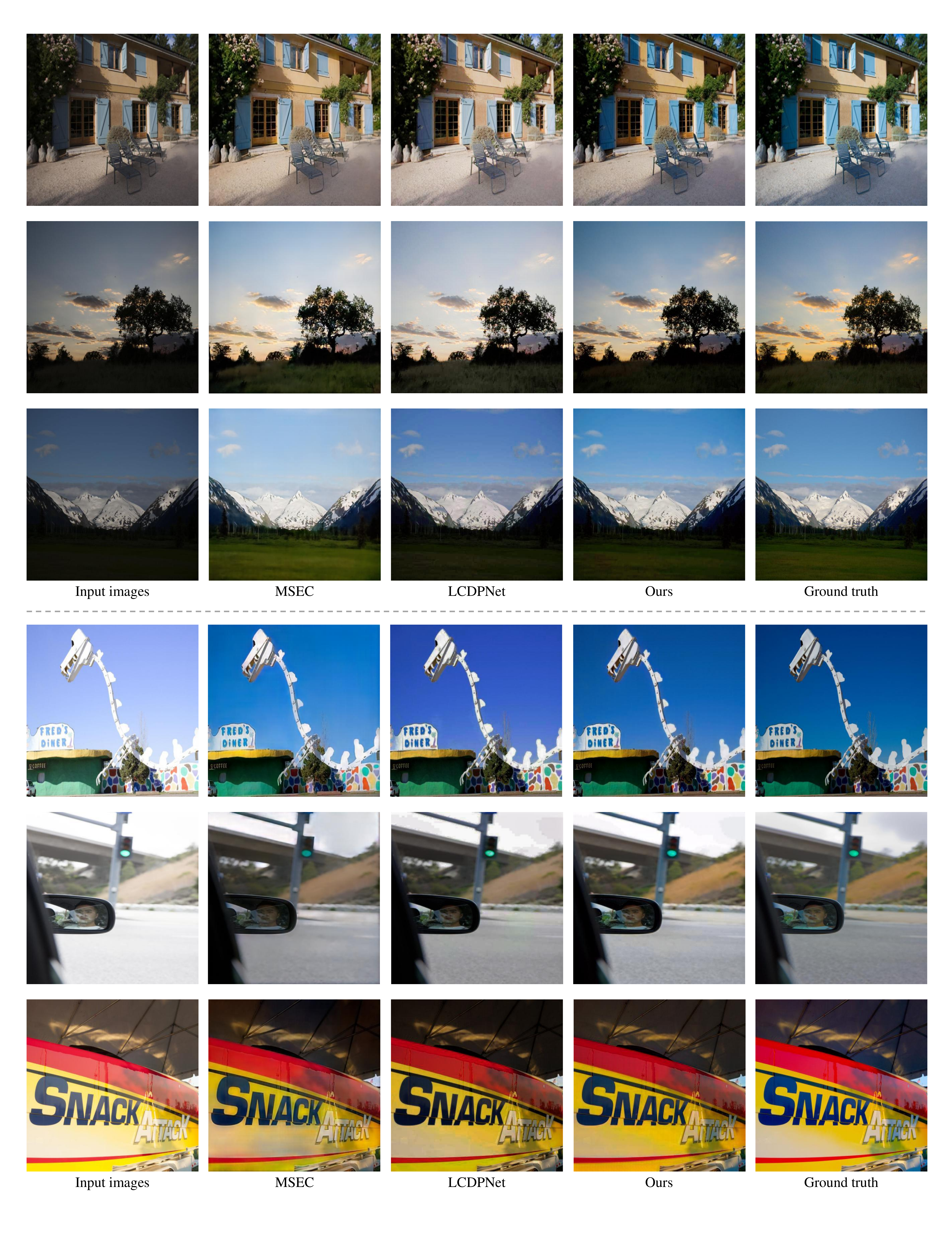} 
}
\caption{Qualitative results. In the presented figures, the top section (1st to 3rd rows) and the bottom section (4th to 6th rows) illustrate comparison results for under-exposed and over-exposed cases, respectively. Compared to MSEC~\cite{afifi2021learning} and LCDPNet~\cite{wang2022lcdp}, our method produces outputs with reduced color bias and fewer visual artifacts. For example, as shown in the 2nd, 3rd, and 4th rows, the sky colors in our results more closely match those in the ground truth images. Moreover, the 5th and 6th rows demonstrate our method's ability to effectively suppress glare in over-exposed scenes, resulting in cleaner outputs with fewer artifacts.
}
\label{fig:visualresult}
\end{figure*}

\paragraph{Quantitative and Qualitative Results} 

We evaluate the performance of our proposed method by comparing it against a broad range of non-learning and learning-based approaches. For learning-based methods not directly evaluated by Afifi et al.~\cite{afifi2021learning} in MSEC, we generate results using their publicly available pre-trained models. For the remaining methods, we adopt the results reported in MSEC~\cite{afifi2021learning}. To assess performance, we use PSNR (Peak Signal-to-Noise Ratio) and SSIM (Structural Similarity Index Measure) as our primary evaluation metrics.

While retouching styles may vary across experts, proper exposure adjustment remains a shared objective. Following the evaluation protocol introduced by Afifi et al.~\cite{afifi2021learning}, we compare our results with all five expert-retouched reference images from the FiveK dataset~\cite{mit5k}. As shown in \Cref{tab:quantresult}, our method achieves the best performance for both over-exposed and under-exposed images, and outperforms existing state-of-the-art approaches on average. \CCH{
Moreover, \Cref{fig:visualresult} illustrates the visually pleasing outputs produced by our model. For instance, the second and third rows in the top section and the fourth row in the bottom section demonstrate that the sky colors in our results are more faithful to the ground truth images.}

\begin{figure*}[t] 
\centering
\includegraphics[width=\textwidth]{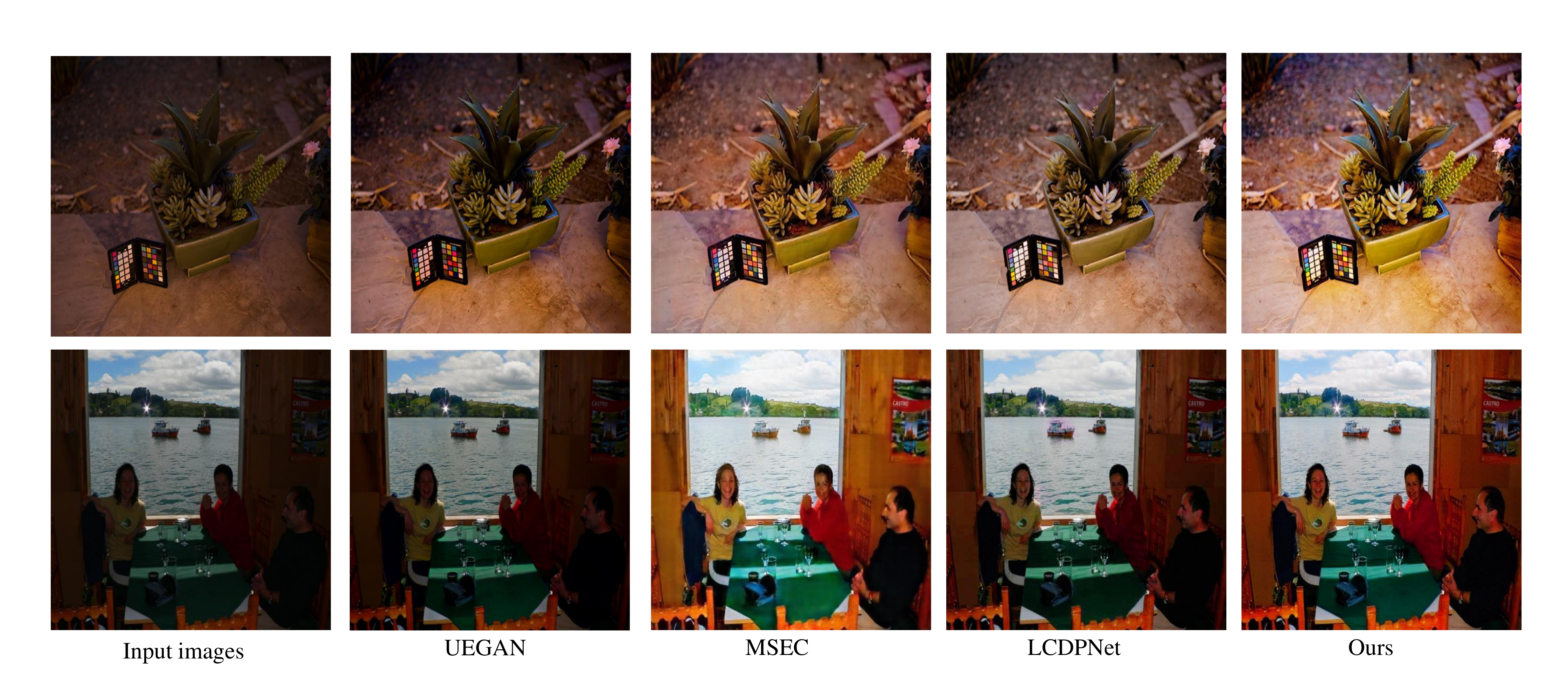}
\caption{Qualitative results on additional real-world datasets for Testing. Despite the absence of ground truth for these datasets, our method's effectiveness is evident even in challenging conditions. The standard color checkerboard presented in the 1st row illustrates our method's capability to reduce color bias effectively without over-enhancement. In the 2nd row, our results showcase the ability to simultaneously enhance indoor and outdoor views, thus demonstrating proficiency in handling high dynamic range imagery. Compared to UEGAN~\cite{ni2020towards} and LCDPNet~\cite{wang2022lcdp}, our approach exhibits superior performance in enhancing indoor details. Furthermore, when contrasted with MSEC~\cite{afifi2021learning}, our method yields a more natural outdoor view, successfully avoiding overexposure issues.
}
\label{fig:result_other}
\end{figure*}
\begin{figure}[htbp]
    \centering
    \includegraphics[width=1.0\linewidth]{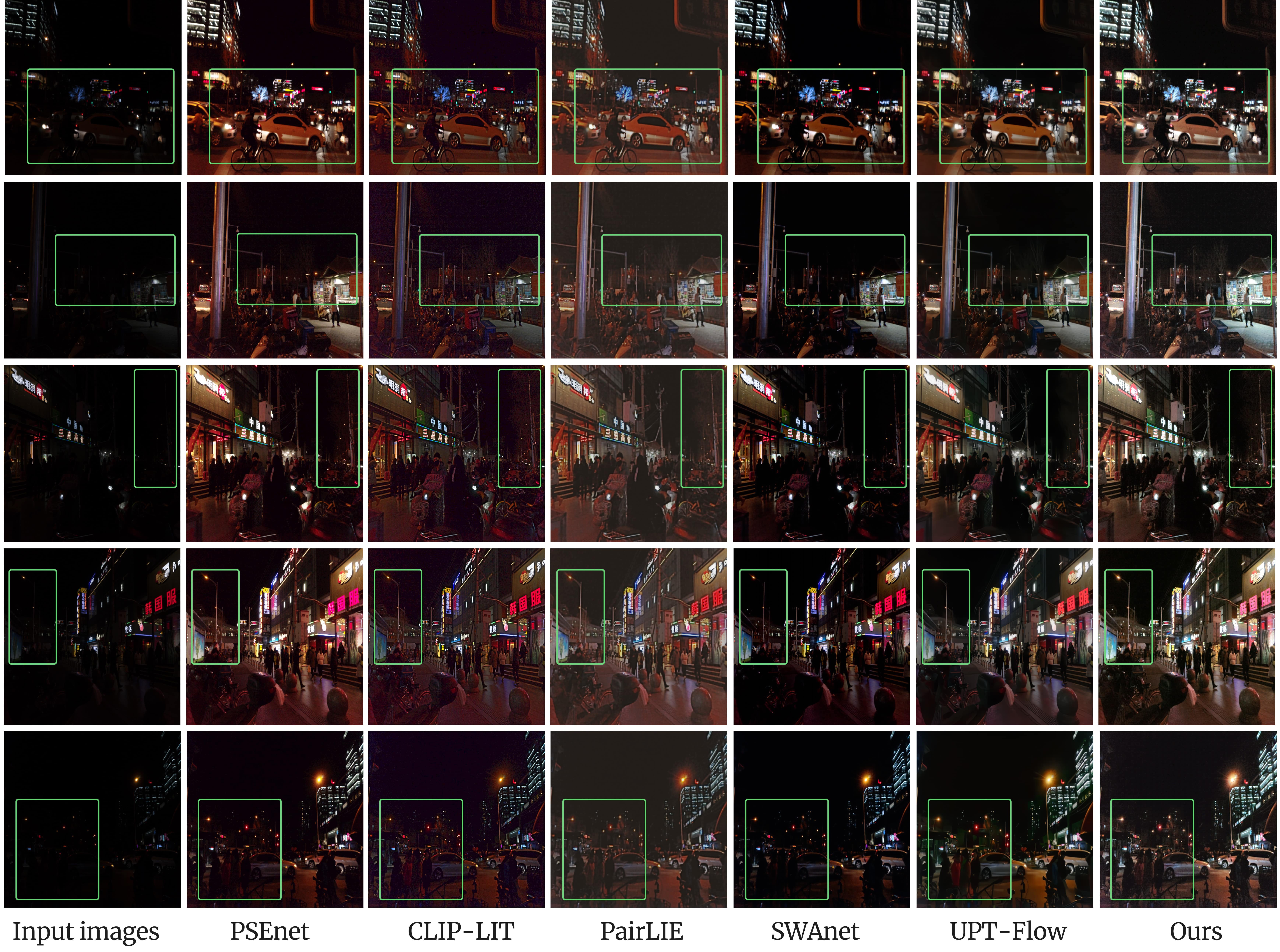} 
   \caption{Qualitative results on challenging nighttime scenes from the DarkFace~\cite{darkface} dataset.}
    \label{fig:darkimage}
\end{figure}

\CCH{
To evaluate the generalization capability of our model, we conduct experiments on several external datasets that cover a wide range of real-world lighting conditions and report the results in \Cref{tab:pi}. As these real-world datasets lack ground truth references, we compute the Perceptual Index (PI)~\cite{pi}, a widely adopted metric for no-reference perceptual quality assessment. The PI is calculated as:

\begin{equation}
    PI=0.5(10-Ma+NIQE),
\end{equation}
where Ma~\cite{ma} and NIQE~\cite{niqe} are two widely used no-reference perceptual quality metrics. Specifically, a higher Ma score and a lower NIQE score indicate better perceptual quality, and consequently, a lower PI value reflects more visually pleasing results. Here, \Cref{tab:pi} presents the perceptual evaluation results on the real-world datasets, including LIME~\cite{lime}, NPE~\cite{npe}, VV~\cite{vv}, and DICM~\cite{dicm}. This evaluation helps assess the effectiveness of our approach in enhancing perceptual quality under diverse real-world lighting conditions, including both underexposed and overexposed regions. As shown in the table, our method outperforms most competing approaches in terms of PI on average. In addition, we provide visual examples from these datasets in \Cref{fig:result_other} to support subjective comparison. The results demonstrate that our enhanced images appear more natural and vivid. 

Finally, to evaluate our model’s ability to recover fine details in extreme dark regions, we first incorporate NLIEE~\cite{nilee}, a no-reference metric specifically designed for low-light image enhancement, to provide a more accurate assessment of our results. We then conduct additional evaluations on challenging low-light benchmarks---DarkFace~\cite{darkface}, MEF~\cite{mef}, and LOL~\cite{lol}---which include nighttime and severely underexposed scenes. As shown in \Cref{tab:nightime}, our method achieves strong performance across both perceptual metrics (NLIEE, PI) and distortion-based metrics (PSNR), demonstrating its effectiveness in enhancing low-light images.

We further present qualitative comparisons in \Cref{fig:darkimage}. It shows that PSENet~\cite{hue2023psenet} and CLIP\_LIT~\cite{clip-lit} introduce a strong red tint that overshadows cooler nighttime hues, while PairLIE~\cite{pairlie} exhibits aggressive contrast enhancement and over-brightening in dark regions. These side effects reduce the naturalness of the final image. SWANet~\cite{swanet} and UPT-Flow~\cite{upt-flow} generate more visually natural results; however, SWANet leaves many shadowed areas overly dim, obscuring important details, and UPT-Flow, though smooth and noise-free, blurs fine textures such as foliage and shop signage. In contrast, our method preserves natural color balance ---retaining both cool and warm tones--- while rendering a realistic soft glow around streetlights and car headlights. At the same time, it effectively restores broad illumination and fine details, such as textures and distant signage, without introducing over-smoothing artifacts.
}

\begin{table}[t]
\centering
\caption{The PI scores on various  real-world lighting image datasets are reported to assess no-reference perceptual quality. Since these datasets lack ground-truth images, reference-based metrics such as PSNR and SSIM cannot be computed; hence, we adopt the PI score for evaluation. Note that lower PI values indicate better perceptual quality. The best results are highlighted in \textbf{bold}, while the second-best results are \underline{underlined}.}
\scalebox{1}{
\begin{tabular}{crrrrr}
\toprule
Method  &  LIME~\cite{lime} & NPE~\cite{npe} & VV~\cite{vv} & DICM~\cite{dicm} & Avg. \\
\midrule
HE~\cite{he}                      & 3.19 & 3.51 & 3.27 & \textbf{3.14} & 3.28  \\
CLAHE~\cite{CLAHE}                & 3.35 & \textbf{3.19} & 3.19 & 3.37 & 3.28  \\
DPE~\cite{Chen:2018:DPE}          & 3.40 & 3.38 & 2.89 & 3.61 & 3.32  \\
UEGAN~\cite{ni2020towards}        & 3.39 & 3.73 & 2.97 & 3.69 & 3.45  \\
Deep UPE~\cite{Wang_2019_CVPR}    & 3.15 & 3.35 & \underline{2.88} & 3.45 & \underline{3.21}  \\
MSEC~\cite{afifi2021learning}         & 3.37 & 3.66 & 2.92 & 3.53 & 3.37  \\
SCI~\cite{ma2022toward}         & 3.21 & 3.27 & 2.99 & 3.45 & 3.23  \\
PSENet~\cite{hue2023psenet}         & \textbf{3.06} & 3.74 & 3.28 & 3.67 & 3.44  \\
IAT~\cite{Cui_2022_BMVC}         & 3.64 & 4.14 & 3.35 & 3.81 & 3.74  \\
LCDPNet~\cite{wang2022lcdp}         & 3.41 & 3.52 & 3.08 & 3.52 & 3.38  \\
UPT-Flow~\cite{upt-flow}         & 3.48 & 3.55 & 3.06 & 3.55 & 3.41  \\
SWAnet ~\cite{swanet}         & 3.03 & 3.46 & 3.03 & 3.55 & 3.27  \\
PairLIE~\cite{pairlie}         & 3.51 & 3.43 & 3.24 & 3.58 & 3.44  \\
Ours                               & \underline{3.11} & \underline{3.26} & \textbf{2.74} & \underline{3.25} & \textbf{3.09}  \\

\bottomrule
\end{tabular}
}
\label{tab:pi}
\end{table}

\begin{table}[ht]
\centering
\caption{Quantitative comparison on challenging nighttime and severely underexposed scenes. For the DarkFace and MEF datasets, we report no-reference metrics (NLIEE and PI), while for the LOL dataset, reference-based metrics (PSNR and SSIM) are used due to the availability of ground-truth images. The best and second performance for each metric is highlighted in \textbf{bold} and \underline{underlined}.}
\begin{tabular}{l|cc|cc|cc}

\hline

Model &  \multicolumn{2}{c|}{DarkFace~\cite{darkface}} & \multicolumn{2}{c|}{MEF~\cite{mef}}  & \multicolumn{2}{c}{LOL~\cite{lol}} \\
 & NLIEE$\uparrow$ & PI$\downarrow$ & NLIEE$\uparrow$ & PI$\downarrow$  & PSNR$\uparrow$ & SSIM$\uparrow$ \\
\hline
PSENet~\cite{hue2023psenet} (2023) &  \underline{41.825} & 2.649 &  \underline{47.919} & 3.469 & 17.461 & 0.650 \\
CLIP\_LIT~\cite{clip-lit} (2023) &  36.833 & \underline{2.575} &  45.148 & \textbf{2.796} & 12.517 & 0.573 \\
PairLIE~\cite{pairlie} (2023) &  40.066 & 3.150 &  45.977 & 3.566  & 18.670 & \underline{0.787} \\
SWANet~\cite{swanet} (2024) &  37.914 & 2.795 &  43.090 & 3.654 & 15.130 & 0.609 \\
UPT-Flow~\cite{upt-flow} (2025) &  41.229 & 3.057  &  44.350 & 3.841 & \underline{18.631} & \textbf{0.854} \\
\texttt{Ours} &  \textbf{41.997} & \textbf{2.350} &  \textbf{48.265} & \underline{3.277} & \textbf{19.501} & 0.759 \\
\hline
\end{tabular}

\label{tab:nightime}
\end{table}

\begin{table}[t]
\centering
\caption{Consistency Evaluation based on the PSNR-Var metric. We also provide the PSNRs of different methods for comparison. Notice that a higher PSNR is better, while a lower PSNR-Var value is better.}
\begin{tabular}{l|cc}
\toprule
\multicolumn{1}{l|}{Method}  &  \multicolumn{2}{c|}{Expert C} \\
\midrule
\multicolumn{1}{l|}{} & PSNR($\uparrow$) & PSNR-Var($\downarrow$) \\
\midrule
DPED (Sony)~\cite{ignatov2017dslr} & 18.02             & 18.23            \\
Deep UPE~\cite{Wang_2019_CVPR}     & 13.69             & 17.28            \\
UEGAN~\cite{ni2020towards} & 16.89 & 12.94 \\
SCI~\cite{ma2022toward}     & 13.31             & 26.94 \\
MSEC~\cite{afifi2021learning}          & 20.21 & \underline{3.24}             \\
PSENet~\cite{hue2023psenet}     & 19.10             & 3.87 \\
IAT~\cite{Cui_2022_BMVC}     & 21.26             & 5.14  \\
LCDPNet~\cite{wang2022lcdp}     & \underline{22.30}             & 4.44 \\
Ours                                & \textbf{22.81}    & \textbf{2.19}    \\

\bottomrule
\end{tabular}
\label{tab:psnrvar}
\end{table}

\begin{table}[t!]
\centering
\caption{The ablation studies on RegNet. The best results are in boldface, and the second-best ones are underlined.}
\scalebox{1.0}{
\begin{tabular}{cccccc}
\toprule
Structures of RegNet  &  \multicolumn{2}{c}{Skip connection} & Cross-attention & PSNR & SSIM \\
\midrule
                   & Path connected & FAM &            &         &          \\
\cmidrule(lr){2-3}
(a) & \XSolidBrush & \XSolidBrush & \XSolidBrush       & 18.11  & 0.584    \\
(b) & \Checkmark   & \XSolidBrush & \XSolidBrush       & 19.29  & 0.752    \\
(c) & \XSolidBrush & \XSolidBrush & \Checkmark         & 18.44  & 0.558    \\
(d) & \Checkmark   & \XSolidBrush & \Checkmark         & \underline{21.96}  & \underline{0.779}    \\
(e) & \Checkmark   & \Checkmark   & \XSolidBrush (FAM) & 19.68  & 0.757    \\
(f) & \Checkmark   & \Checkmark   & \Checkmark         & \textbf{22.05}  & \textbf{0.788}    \\
\bottomrule
\end{tabular}
}
\label{tab:componenteffec}
\end{table}

\paragraph{Consistency Evaluation} To understand the ability of a method to produce consistent outputs under diverse exposure settings, we calculate the variance of PSNR values of exposure-corrected images with the same scene but different exposure values (EVs=-1.5, -1, 0, 1, 1.5). We denote this consistency evaluation metric as ``PSNR-Var''. A lower PSNR-Var value means that the approach to be evaluated is more stable. \Cref{tab:psnrvar} shows the consistency evaluation and comparison. In this experiment, the PSNR values used for calculating PSNR-Var are calculated using the Expert C dataset as ground truth. We can see that our method produces a lower PSNR-Var. Similar observations can also be found in \Cref{tab:quantresult}. For example, using Expert C as the ground truth, the average PSNR of our method is \CCHR{22.89dB} for the over-exposed case and \CCHR{22.70dB} for the under-exposed case. Thus, the PSNR difference of our approach is \CCHR{0.19dB}. In contrast, the method in ~\cite{afifi2021learning}, also proposed for both over- and under-exposure correction, produces a \CCHR{0.56dB} PSNR difference. If we check the method designed only for under-exposure correction, for instance, the method Deep UPE~\cite{Wang_2019_CVPR}, 
the PSNR difference goes highly to 8.45dB.

\CCH{
\paragraph{Component-wise Ablation Study}
To thoroughly evaluate the effectiveness of each stage, we conduct a component-wise ablation study by replacing either MEGNet with three traditional exposure generation methods ---Gamma Correction~\cite{gammacorrection}, Log Transform~\cite{logtransform}, and Brightness Scaling--- or replacing RegNet with Exposure Fusion~\cite{exposurefusion} as the representative traditional fusion technique. This study enables us to isolate and quantify the contribution of each component toward the final restoration performance.
As presented in \Cref{tab:traditional}, the results reveal that MEGNet surpasses traditional exposure generation methods even when combined with a conventional fusion approach, achieving 16.493 PSNR and 0.771 SSIM. This is because MEGNet is trained to simulate realistic exposure variations using Adobe Camera Raw, whereas traditional methods rely on simple approximations. Similarly, RegNet surpasses Exposure Fusion~\cite{exposurefusion} across all exposure inputs, with the MEGNet–RegNet combination yielding the best outcome (22.810 PSNR and 0.860 SSIM). Although traditional exposure generation methods also benefit from the RegNet, their performance remains consistently lower, confirming the importance of both proposed modules and their synergy in delivering superior enhancement quality.
}
\begin{table}[ht]
\centering
\caption{Component-wise ablation study on Expert C dataset. }
\begin{tabular}{|l|l|c|c|}
\hline
\textbf{Exposure Generation Method} & \textbf{Fusion Method} & \textbf{PSNR $\uparrow$} & \textbf{SSIM $\uparrow$} \\
\hline
Gamma Correction~\cite{gammacorrection}  & \multirow{4}{*}{Exposure Fusion~\cite{exposurefusion}} & 11.771 & 0.543 \\
Log Transform~\cite{logtransform}     &                        & 13.647 & 0.701 \\
Scale Adjustment  &                        & 16.431 & 0.748 \\
MEGNet (Ours)        &                        & \textbf{16.493} & \textbf{0.771} \\
\hline
Gamma Correction~\cite{gammacorrection}  & \multirow{4}{*}{RegNet (Ours)}  & 16.699 & 0.735 \\
Log Transform~\cite{logtransform}     &                        & 17.345 & 0.663 \\
Scale Adjustment  &                        & 14.080 & 0.729 \\
MEGNet (Ours)        &                        & \textbf{22.810} & \textbf{0.860} \\
\hline
\end{tabular}
\label{tab:traditional}
\end{table}

\subsection{Ablation Studies on RegNet}
\label{sec:AblationRegNet}


In this section, we present ablation studies to thoroughly analyze the effectiveness of each component within RegNet. In the experimentation, we utilized a patch size of $512\times512$ for these ablation studies on RegNet. Below, we detail the comparison of six different settings, as outlined in \Cref{tab:componenteffec}:

\begin{enumerate}
\item[(a)] Only the four-layer convolutional Encoder-Decoder network is applied in RegNet.
\item[(b)] Only the four-layer convolutional Encoder-Decoder network with skip connection is applied in RegNet.
\item[(c)] The four-layer convolutional Encoder-Decoder network and the cross attention regressor are applied in RegNet.
\item[(d)] The four-layer convolutional Encoder-Decoder network with skip connection and the cross attention regressor are applied in RegNet.
\item[(e)] The four-layer convolutional Encoder-Decoder network with FAM in the skip connection path is applied in RegNet. Besides, we substitute FAM for cross attention for comparison.
\item[(f)] Fully-equipped RegNet.
\end{enumerate}

Note that (a) and (b) only keep the four-layer convolutional encoder-decoder in RegNet while (e) replaces all the cross-attention with FAM. Besides, the “path connected” column means there exists a skip connection between the encoder and decoder or not, while the “FAM” column denotes whether FAM exists in this path.
As shown in \Cref{tab:componenteffec}, by comparing setting (c) with setting (a), where the PSNRs are 18.435 dB vs. 18.106 dB, as well as setting (d) with setting (b), where the PSNRs are 21.957 dB vs. 19.291 dB, we can conclude that using the cross-attention module in the regressor can achieve better performance than the settings without it. Moreover, we also demonstrate the effectiveness of the FAM module by removing it from the skip connection path. Setting (f) and (d) show that PSNR drops around 0.1 dB without FAM. On the other hand, in setting (e), we try to replace the cross-attention module with FAM since FAM can also model the relationship between the input features and optimal exposure values to produce adjusted output features. However, comparing the results of (e) and (f), the cross-attention-based regression remarkably outperforms the FAM version. The reason is that cross-attention refers to the full features for exposure adjustment, while FAM only leverages local information.

\begin{figure*}[t] 
\centering
\includegraphics[width=\textwidth]{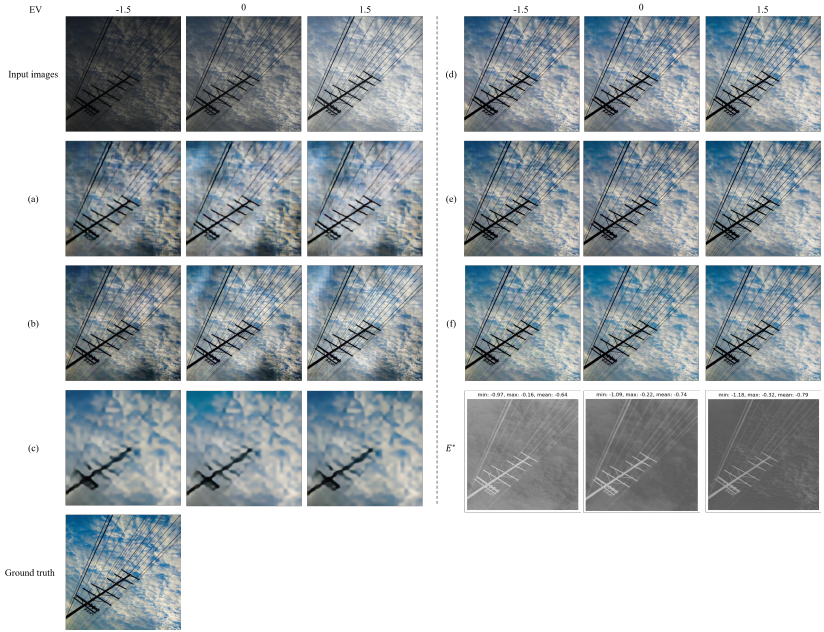}
\caption{Qualitative comparison of the ablation studies on RegNet. \textcolor{black}{(a)-(f) correspond to the variant structures defined in \Cref{tab:componenteffec}.}}
\label{fig:componenteffec}
\end{figure*}

A qualitative comparison is also provided in \Cref{fig:componenteffec}. Our final version (i.e., the setting (f)) generates more consistent exposure-corrected outputs for the testing photos with the same content but different exposures. If checking deeper, we may find that the results of settings (c), (d), and (f) have consistent predictions over different exposures. It is because the settings are all realized under the regression framework for image exposure enhancement. It proves that the regression framework is critical in yielding physically accurate/consistent correction and makes our ExReg different from the other related works.

Next, we visualize the exposure map of setting (c) (at the bottom right of \Cref{fig:componenteffec}). The map shows our ability to determine the correct exposure for local regions and reveals the adaptivity of the cross-attention module and the exposure predictor without using FAM. Finally, from (e) and (f), we see that FAM in the skip connection path can help to provide more image details and make the generated images more natural.

\subsection{Experiments on MEGNet}
\label{sec:ExpRegNet}
In this section, two in-depth analyses of MEGNet are provided. We start to demonstrate the effect of  $\Delta EV$s on \cite{afifi2021learning} and show the appropriate range for the image synthesis task. Then, in the following subsection, we explore the impact of the number of generated images on the regression-based exposure correction. \CCHR {In these experimentations, we utilized a patch size of $512\times512$ for analysis.}

\begin{table}[t]
\centering
\caption{Quantitative evaluation of the MEGNet's image generation ability under different $|\Delta EV|$}
\begin{tabular}{crr}
\toprule
$|\Delta EV|$  &  PSNR & SSIM \\
\midrule
3.0     & 29.12  & 0.936    \\
2.5     & 30.12  & 0.943    \\
2.0     & 31.31  & 0.948    \\
1.5     & 33.01  & 0.956    \\
1.0     & 34.74  & 0.963    \\
0.5     & 37.45  & 0.970    \\
0       & 42.90  & 0.988    \\
\bottomrule
\end{tabular}

\label{tab:meg}
\end{table}

\begin{figure}[!t] 
\centering
\includegraphics[width=0.45\textwidth]{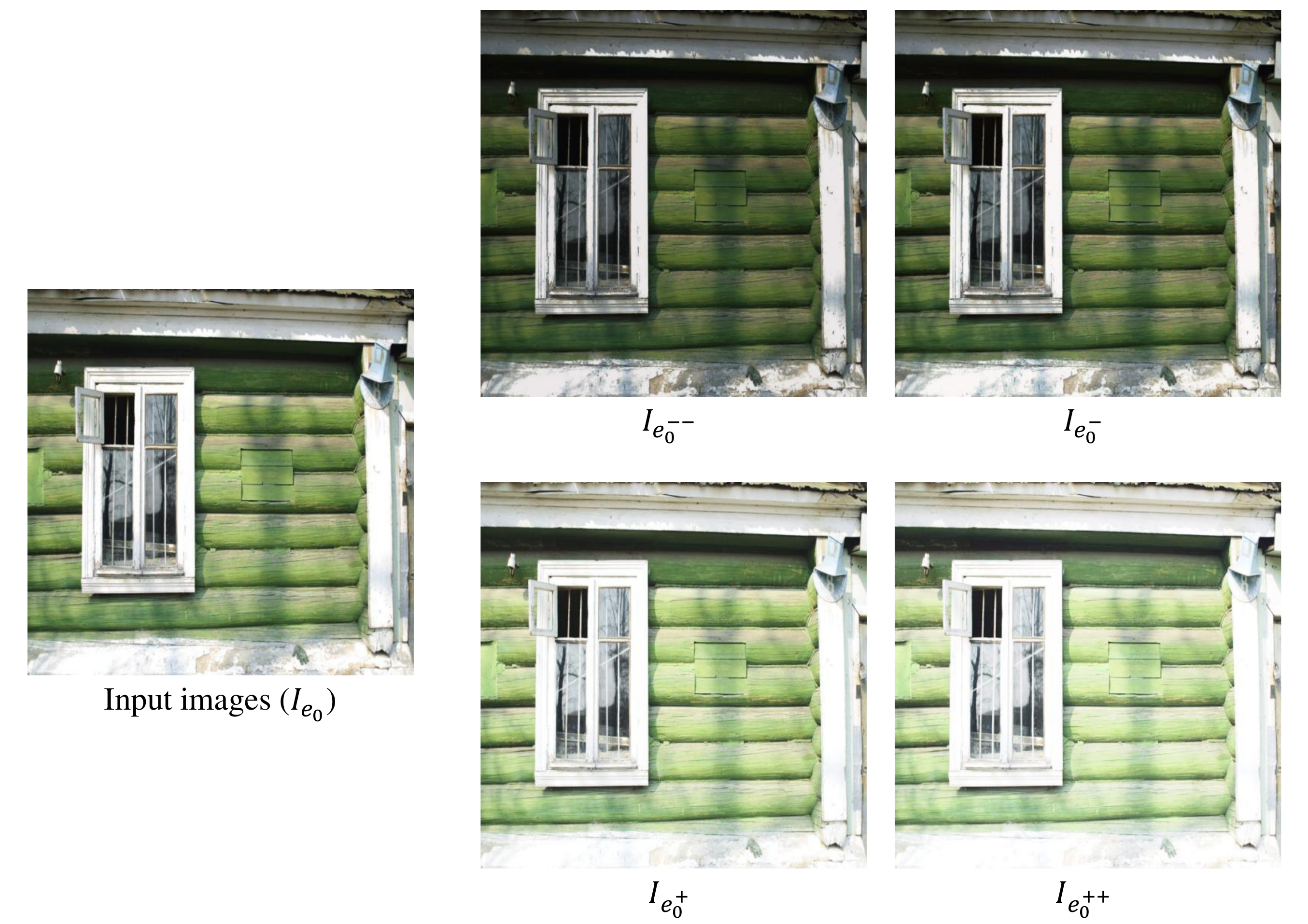} 
\caption{Synthetic images ($I_{e_0^{--}}, I_{e_0^-}, I_{e_0^+}, I_{e_0^{++}}$) generated by our MEGNet given the input image ($I_e$) and $\Delta EV$ = -1.5, -1.0, 1.0, and 1.5.}
\label{fig:meg}
\end{figure}

\paragraph{The effect of $\Delta EV$ }
This evaluation is carried out on the dataset provided by ~\cite{afifi2021learning}, which consists of 5,905 images rendered with different exposure settings. Specifically, the dataset contains 1,181 images and their derivatives with four relative EVs ($\Delta EV$ = -1.5, -1.0, 1.0, 1.5). For evaluation, we generate 5 images to approximate the true images with EVs equal to ${e-1.5, e-1.0, e, e+1.0, e+1.5}$ by feeding a reference image $I_e$ and $\Delta EV$ into our MEGNet network. Here, the case, $\Delta EV$ = 0, is included to validate our fidelity, and the other four generated images are denoted as  $I_{e_0^{--}}$, $I_{e_0^-}$,  $I_{e_0^{}}$,  $I_{e_0^+}$, and $I_{e_0^{++}}$. Referring to the ground truth in the dataset, \Cref{tab:meg} reports the average PSNR and SSIM results of the generated images under different $\lvert\Delta EV\rvert$ (i.e., the ground truth image and the generated image, chosen for evaluation, have $\lvert\Delta EV\rvert$ exposure difference). The table shows that MEGNet can generate images with good quality when  $\lvert\Delta EV\rvert\leq 1.5$. Moreover, the smaller the absolute value of $\Delta EV$, the better the quality of the generated image. That is why we choose $\lvert\Delta EV\rvert\leq 1.5$ in our system. We also visualize one set of generation results for $I_{e_0^{--}}$, $I_{e_0^-}$, $I_{e_0^+}$, and $I_{e_0^{++}}$ given its input image $I_{e_0}$ as reference in \Cref{fig:meg}. It shows that MEGNet can generate vivid images under different exposure settings with few artifacts.

\begin{table}[t]
\centering
\caption{Quantitative analysis of the impact of the number ($N$) of generated images on the regression-based exposure correction.}
\begin{tabular}{crr}
\toprule
N  &  PSNR & SSIM \\
\midrule
2     & 21.81  & 0.785    \\
4     & 22.05  & 0.788    \\
6     & 22.02  & 0.789    \\
\bottomrule
\end{tabular}

\label{tab:numberN}
\end{table}

\begin{figure}[t] 
\centering
\includegraphics[width=0.45\textwidth]{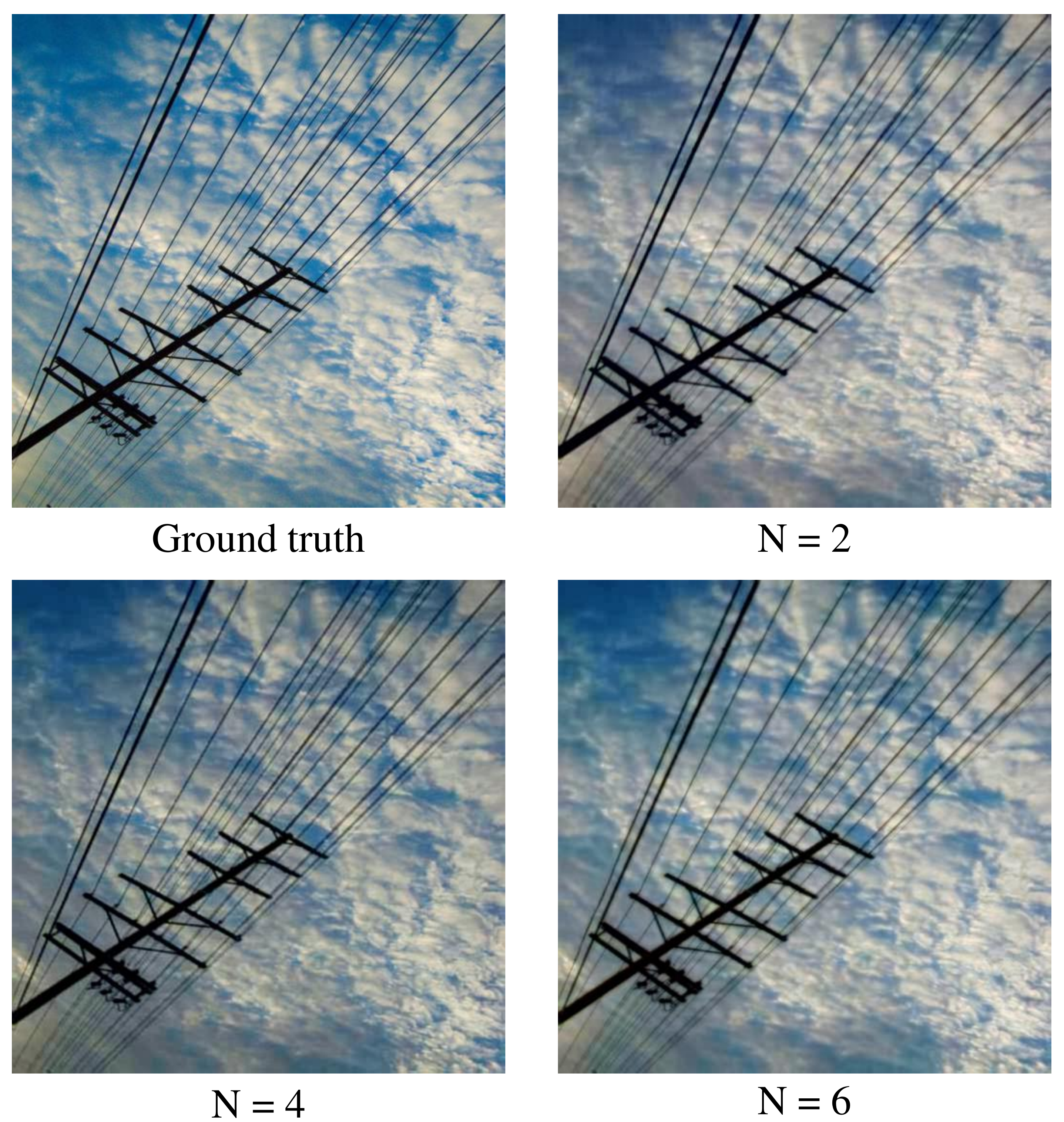} 
\caption{Exposure-corrected results under different numbers of generated images. $N$ is the image number.}
\label{fig:numberN}
\vspace{-2mm}
\end{figure}

\paragraph{The effect of the number of generated images}
To explore how the number of the generated images (denoted as $N$) affects the correction performance, we train three networks with $N$ = 2, 4, and 6, respectively. Besides, within the range $\lvert\Delta EV\rvert\leq 1.5$ where MEGNet works well, the generated images that can cover a broader range of exposure settings are preferred for the regressor to achieve high quality. Coincidentally, we find the authors in ~\cite{afifi2021learning} suggest generating their exposure dataset containing images with EV equal to -1.5, -1.0, 0, 1.0, and 1.5. Therefore, we choose $\Delta EV$ = -1.5 and 1.5 for the model with $N$ = 2, $\Delta EV$ = -1.5, -1.0, 1.0, 1.5 for the case with $N$ = 4, and $\Delta EV$ = -1.5, -1.0, -0.5, 0.5, 1.0, 1.5 for the one with $N$ = 6. As shown in \Cref{tab:numberN} and \Cref{fig:numberN}, more generated images would yield better correction results owing to more accurate regression. However, a greater N would infer more computational complexity. To strike a balance, we choose $N$ = 4 as our setting for its comparability with $N$ = 6.

\section{Conclusion}
\CCH{
This work presents a novel framework, ExReg, for wide-range photo exposure correction. We formulate the exposure correction task as a multi-dimension regression process and propose a novel two-stage approach. In the first stage, we design MEGNet to generate images with variant exposure settings for multi-dimension regression in the next step. Next, RegNet, built upon Encoder-ANP-decoder and including an auxiliary module for image details preservation, is applied to predict the region-wise exposure values and generate the final well-exposed image. Our ExReg, combining MEGNet and RegNet, outperforms SOTA algorithms in terms of PSNR and SSIM and can yield more visually consistent results given variant testing images under diverse exposures.}

\section{Acknowledgments}{

This work was financially supported in part (project number: 112UA10019) by the Co-creation Platform of the Industry Academia Innovation School, NYCU, under the framework of the National Key Fields Industry-University Cooperation and Skilled Personnel Training Act, from the Ministry of Education (MOE) and industry partners in Taiwan. It also supported in part by the National Science and Technology Council, Taiwan, under Grant NSTC-114-2218-E-A49-024-, Grant NSTC-112-2221-E-A49-089-MY3, Grant NSTC-114-2425-H-A49-001, Grant NSTC-113-2634-F-A49-007, Grant NSTC-112-2221-E-A49-092-MY3, and in part by the Higher Education Sprout Project of the National Yang Ming Chiao Tung University and the Ministry of Education (MOE), Taiwan. It is also partly supported by MediaTek Inc., Hon Hai Research Institute, and Industrial Technology Research Institute. The authors would like to express their sincere gratitude to Prof. Ching-Chun Huang for his valuable guidance and continuous support throughout this research.

}
\bibliographystyle{ACM-Reference-Format}
\bibliography{sample-base}

\end{document}